\documentclass{article} 

\usepackage{graphicx}
\newcommand{\figref}[1]{Fig.~\ref{fig:#1}}

\usepackage{amsmath}
\usepackage{hyperref}

\newcommand\norm[1]{\left\lVert#1\right\rVert}
\newcommand\dwmin{\Delta w_\text{min}}

\begin{document}

\title{Training large-scale ANNs on simulated resistive crossbar
  arrays}

\author{Malte J. Rasch\,$^{*}$, Tayfun Gokmen, \and Wilfried
  Haensch\\ {\small IBM Research AI, TJ Watson Research Center,
    Yorktown Heights , New York, USA}}

\maketitle

\begin{abstract}
  Accelerating training of artificial neural networks (ANN) with
  analog resistive crossbar arrays is a promising idea. While the
  concept has been verified on very small ANNs and toy data sets (such
  as MNIST), more realistically sized ANNs and datasets have not yet
  been tackled.  However, it is to be expected that device materials
  and hardware design constraints, such as noisy computations, finite
  number of resistive states of the device materials, saturating
  weight and activation ranges, and limited precision of
  analog-to-digital converters, will cause significant challenges to
  the successful training of state-of-the-art ANNs. By using analog
  hardware aware ANN training simulations, we here explore a number of
  simple algorithmic compensatory measures to cope with analog noise
  and limited weight and output ranges and resolutions, that
  dramatically improve the simulated training performances on RPU
  arrays on intermediately to large-scale ANNs.
\end{abstract}

\section{Introduction}

The amount of computation needed to train modern deep learning
networks is immense. Recently, it has been suggested to use resistive
crossbar arrays to accelerate parts of the ANN training in analog
space, with a potential dramatic increase in computational performance
compared to digital
systems~\cite{yang2013memristive,burr2017neuromorphic,fumarola2016accelerating,ambrogio2018equivalent,tg_dnn,haensch2019next}. While
analog crossbar arrays, also termed resistive processing units (RPU)
arrays~\cite{tg_dnn}, could speedup inference of DNNs (deep neural
networks) and CNNs (convolutional neural
networks)~\cite{shafiee2016isaac,yakopcic2017extremely}, the true
benefit of an analog deep learning accelerator lies in the
acceleration of the training process as well, since training of ANNs
is generally orders of magnitude more computationally demanding than
inference. However, many significant design, material, and algorithmic
challenges still need to be addressed in order to enable training on
RPUs with high accuracy.

While forward and backward pass of the stochastic gradient descent
(SGD) are relatively straightforward to implement in analog hardware,
a truly in-memory weight update that matches the performance of
backward and forward passes in computing a pass in constant time, is
much more challenging. A fully parallized update is important,
however, because if the gradients were instead be computed in the
digital part of the system, it would require on the order of $n^2$
computations and thus all speed advantages of the analog
(i.e. computing the forward and backward in constant time), would be
lost.
  
One promising design is to use stochastic pulse sequences to
incrementally update the weight elements in a parallel
fashion~\cite{tg_dnn}.  This approach was explored for small to
moderate DNNs and CNNs\cite{tg_dnn, tg_cnn, rasch2018efficient}, as
well as LSTMs~\cite{tg_lstm} in simulations. It was shown that
moderate amounts of analog noise and physically limited weight update
sizes can be tolerated during training, in particular, if one
introduces additional noise and bound management techniques, that
mitigate the analog noise in the backward pass and can be computed in
linear time.

It was also noticed, however, that it is critical for the SGD
algorithm to employ device materials that have symmetric switching
behavior~\cite{tg_dnn}, in other words, a single update pulse in
positive or negative direction should effectively change the weight by
a similar amount at least on average. To achieve this balanced update
behavior requires significant efforts on the material development for
RPUs or significant changes in gradient decent algorithms or
network architectures.
  
Another problem is the limited weight ranges and limited number of
states supported by the device material, which, in floating point
terms, is related to the bit resolution of the weights.  Although the
resistance of the memristive device can be set to any analog value in
principle, materials are inherently subject to noise, and hardware
designs require that the weight update is pulsed, where each pulse
will increase or decrease the weight value by an finite amount of
$dwmin$ on average. Thus, if the weight is bounded in the
range $w \in [-w_b,w_b]$, the number of states can be defined as
$N_s \equiv \frac{2w_b}{\dwmin}$. Note, however, that
cycle-to-cycle variation of the weight update is generally large and
the read out process is noisy, too, so that a single read might not be
able to discern neighboring states.

In this paper, we focus on the latter issues, that is the limited
number of material states, restricted weight ranges, and the noise
in the training process. We here ask the following questions: even if
we had a device with noisy but (on average) ideally symmetric
switching behavior, how many states are necessary to successfully
train larger scale models on more challenging image data sets (than
MNIST)?  Are there simple ways to improve performance on RPU arrays
with their limited weight resources?

Our contribution are: (1) Scaling up the hardware realistic
pulsed-update training on simulated RPUs to larger networks with $>60$
million weights and $>1.2$ million images, such as AlexNet on the
ImageNet data set~\cite{alexnet}, (2) showing the importance of
normalization to overcome analog noise, (3) devising a new scheme for
virtually remapping the weight ranges to maximize usable resistive
states and achieving proper regularization on RPUs arrays.

\begin{figure}[t]
  \centering
  \includegraphics[width=\textwidth]{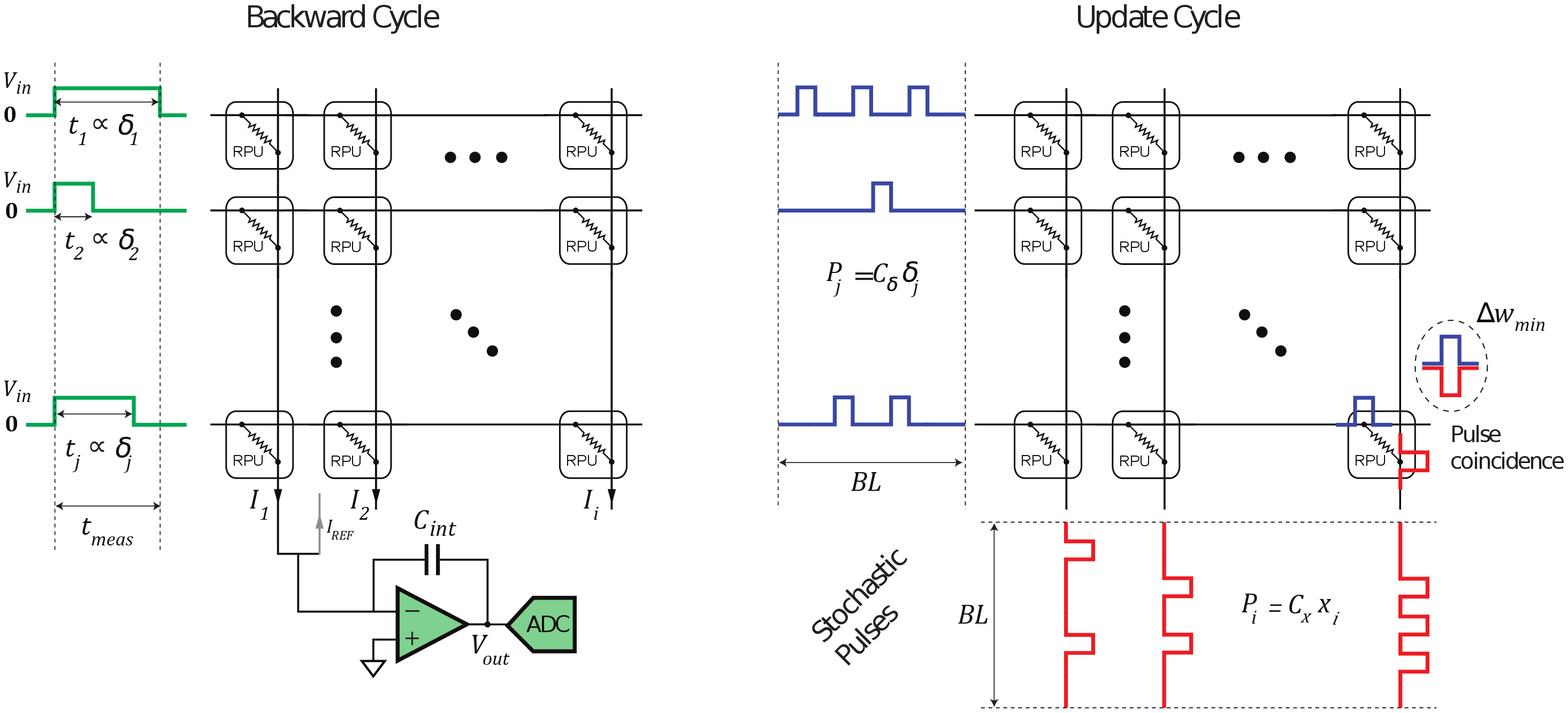}     
  \caption{\small Backward pass and pulsed update for RPU arrays
    hardware implementation, adopted from~\cite{tg_dnn}. Forward pass
    is analogous to the backward pass (with input and output
    reversed). (Left) Matrix-vector operation are performed in analog
    space, where memristors encode the weights of the matrix and
    inputs are encoded temporally with voltage pulses of variable
    length. (Right) During update cycle, stochastic pulse trains are
    generated, where the probability of pulse occurrence encodes the
    input values.  If pulses from both input sides coincide at a RPU
    device, its resistive value is updated corresponding to (on
    average) $\dwmin$ in weight units.  }
  \label{fig:RPU}
\end{figure}

\section{RPU model and simulations}

Our simulation is based on the RPU model proposed by \cite{tg_dnn}
(see \figref{RPU} for an illustration). We, however, adapted the
previous C++-simulator of \cite{tg_dnn} to integrate with the Caffe2
machine learning
framework\footnote{\url{https://caffe2.ai/}}~\cite{caffe}, to be able
to flexible handle different network architectures and datasets. We
also re-implemented the RPU simulation code to fully support GPUs
acceleration to improve the runtime for larger models and convolutions
for inference and training. For the ConvNets investigated below, a
training simulation with fully pulsed update typically only runs 2-3
times slower than a native floating point training in Caffe2.

We adopt the scheme of \cite{tg_cnn} and use stochastic pulse trains
of maximal length 31. We ensured in our simulation that pulsed weight
update is done (1) by drawing actual stochastic pulse trains for each
update\footnote{We, however, used only 2 instead of the 4 positive and
  negative pulse train combinations during update for speed
  advantages. We did not notice any different behaviour by this slight
  simplification.} and calculating the coincidence of pulses per
weight, (2) for every coincident pulse occurrence the corresponding
device weight (conductance) is updated by a single step drawn from a
Gaussian distribution (with standard deviation 30\% of the mean
$\dwmin$) and saturating the hard bounds if necessary, (3) the
sequence order of updates (in case of a batch learning or
convolutions) is preserved as if would be done in hardware.

In our simulations, all analog noise, such as circuit components and
peripheral noise, is referred to the output of the analog computation
and modeled as Gaussian noise processes added to each analog output
line. The noise values are re-drawn for each analog
computing step, i.e. each computed vector-matrix product. The device
specification in \cite{tg_dnn} gives reasons to set the standard
deviation of these cumulative noise terms to $0.06$.

\begin{figure}[t]
  \centering
  \includegraphics[width=\textwidth]{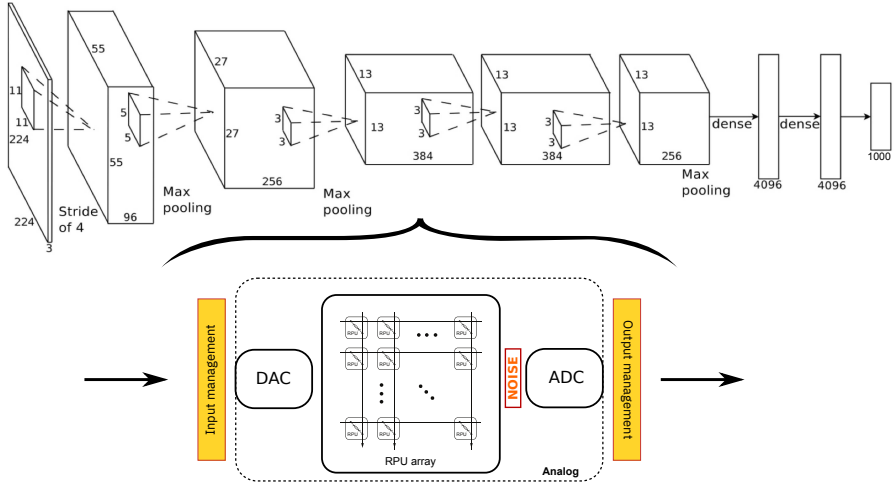}     
  \caption{\small In the RPU simulations Caffe2 fully-connected and
    convolution layers are replaced by operators that contain the
    analog hardware simulation. Given a network architecture,
    e.g. AlexNet~\cite{alexnet} in this illustration, each layer will
    be translated to a separate RPU operator (of matching sizes of the
    weight matrices). The RPU operator contains input and output
    management in the digital part, where inputs are normalized or
    scaled by the maximal input value in floating point precision, as
    described in the text. DAC and ADC have limited ranges and
    precision and the analog noise is drawn each operation cycle from
    a Gaussian distribution and added to each analog output line (see
    \cite{tg_dnn} for details of our RPU model).}
  \label{fig:sim}
\end{figure}

Additionally, we assume that each analog RPU array stores the weights
of a layer (the kernel matrix) and performs matrix vector products in
a way described in \cite{tg_dnn,tg_cnn}. The RPU array is
communicating with the next layer in digital space, thus we assume
analog-to-digital (ADC) and digital-to-analog (DAC) converters per RPU
array~(see \figref{sim}). The DAC/ADC discretizes the input values
into $m$ bins in the range of its bounds (which are fixed by hardware
design to $\pm 1$ for DAC, and $\pm 12$ for the ADC, see our
definition of the baseline RPU model~\cite{tg_cnn}). The bit
resolution of the converters are then $\log_2 m$. We here assume 7~bit
for DAC and 9~bit for ADC if not stated otherwise  (see also
\cite{tg_lstm} for a discussion).

For the SGD training, we just use plain batch-wise SGD without any
additional momentum or weight decay, which would be difficult to
efficiently implement on analog hardware architectures.

\subsection{Noise, bound, and update management techniques}
It has been shown previously~\cite{tg_cnn} that it is essential to
introduce noise management techniques on the digital side, to cope
with the noisy analog computations as well as the bounded ranges of
the inputs and outputs of the analog RPU. 

Noise management becomes vitally important during backward pass, since
the backward propagated errors are usually orders of magnitude smaller
than the forward pass values and would be buried in the analog noise
floor if not properly re-scaled. In particular, we use the noise
management introduced by~\cite{tg_cnn}, where the digital input vector
$\mathbf{x}$ is divided by $\alpha=\max |x_i|$ before the DAC and
re-scaled again by $\alpha$ after the ADC in digital. Additionally, we
use a bound management (only in the forward pass), that iteratively
multiplies $\alpha$ by factors of 2 until the ADC bound does not
saturate any of the outputs anymore. In this way, larger output values
can be accommodated, with the cost of ADC resolution (which is
effectively reduced by a factor of 2 for each iteration) and cost of
runtime since the computation of the forward pass is essentially
repeated multiple times. However, since one cycle is very fast (order
of 100ns) and the geometric reduction of $\alpha$ does not need many
iterations (at the very most the number of bits of the ADC), the
additional runtime cost seems tolerable if necessary for high
accuracy, which is in particular important before the softmax
layer. Also, hardware solutions could cut short the integration time
further, by triggering an abort when one output saturates early.

In addition, we use an update management introduced in~\cite{tg_cnn},
that rescales the pulse generation probabilities based on the
$\max |x_i|$ and $\max |d_j|$ and the expected pulse width $\dwmin$,
where $\mathbf{d}$ denotes the error vector during update. We refer to
\cite{tg_cnn} for the details.

\section{Results}

We first investigated a small 3-layer convolutional network plus one
fully-connected layer and ReLu activation\footnote{We used the
  ``Full'' network (except from changing the sigmoid activations to
  ReLu) from the
  \href{https://github.com/BVLC/caffe/tree/master/examples/cifar10}{Caffe
    cifar10 examples}.}  on the CIFAR10 data set (with weak data
augmentation, e.g. mirroring and color jittering). The network has 79328
weights and uses lateral response normalization (LRN) between
convolution layers. The image size is $32\times 32$ pixels.  While of
similar model and data size as the CNN previously investigated on the
MNIST data set~\cite{tg_cnn}, the CIFAR-10 dataset contains rescaled
color images, whose classification is much more challenging than
classifying the cleanly handwritten binary digits of MNIST: Using
floating point (FP) and no input data augmentation, the above CNN
achieves about 0.8\% test error on MNIST, while only 25\% test error
on CIFAR-10.

\begin{figure}[t]
  \centering
  \includegraphics[width=\linewidth,clip,trim=2cm 0 2cm 0]{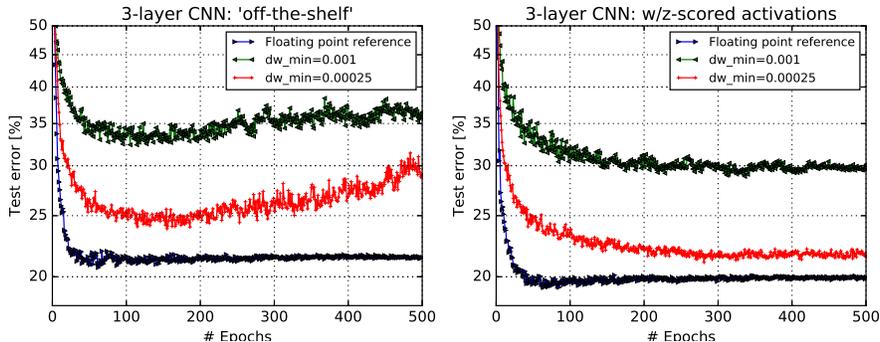}
  \caption{\small 3-layer CNN network on CIFAR10 trained with the baseline RPU
    model~\cite{tg_cnn}, except using balanced switching
    behavior. (Left) Using the 'off-the-shelf' model with the RPU
    simulation for each layer, performance is severely impaired
    compared to the floating point reference, even if the number of
    states is increased $4\times$ (red line). (Right) Normalization of
    the (digital) activations between layers improves RPU performance
    and eliminates overfitting (ie. late rise in test error, see
    left). Parameters: $\lambda=0.1$, multiplied by 0.8 every 20
    epochs; no weight decay; batch size 100.} 
  \label{fig:full}
\end{figure}

We first trained the network with our baseline RPU model (see
\cite{tg_cnn} for definition), except with balanced weight update and
7 bit DAC and 9 bit DAC resolutions. This RPU model has 1200 weight
states on average per device ($\dwmin=0.001$ and $w_b=0.6$) and
gave very acceptable performance for a smaller CNN on MNIST (compare
to \cite{tg_cnn}~Figure 4, ``All no imbalance''). Nevertheless, here
we found that performance is dramatically impaired, see \figref{full}
(left, green curve), albeit using the same bound and noise management
techniques described in~\cite{tg_cnn}. Even increasing the number of
available states by $4\times$ does not reach FP performance
(\figref{full}, left, red curve).

Thus, more challenging data sets and larger networks
seem to again require additional algorithmic improvements for training
RPUs. In the following, we introduce two simple remedies.

\subsection{Normalization balances activations in the presence of
  noise}

The reason for the poor learning ability becomes clear when estimating
the signal-to-noise ratio of the analog matrix product. For each
analog output, we have $y_i = \mathbf{w}_i\mathbf{x} + \xi$, where
$\mathbf{w}_i$ is the $i$th row of the analog weight matrix and $\xi$
the analog noise term, i.e. a Gaussian process with zero mean and
standard deviation $\sigma$. If we assume that $\mathbf{w}_i$ is a
``good'' feature vector for input $\mathbf{x}$, the direction of
$\mathbf{x}$ and $\mathbf{w}_i$ should approximately match, thus
$\mathbf{w}_i\approx
\frac{\norm{\mathbf{w}_i}}{\norm{\mathbf{x}}}\mathbf{x}$. Thus for a
number of similarly well matching inputs, the signal-to-noise ratio
$\frac{\langle y_i^2\rangle}{\sigma^2}$ is roughly
\begin{equation}
  \label{eq:snr}
  \text{SNR} \propto \frac{\norm{\mathbf{w}_i}^2 \langle \norm{\mathbf{x}}^2\rangle}{\sigma^2}.
\end{equation}
Although this is only an approximate calculation, it shows that weight
vectors, that are well matched with the inputs, will quickly grow in
norm to improve the signal-to-noise ratio during initial SGD training.

Moreover, since only few rows of $W$ matches the input initially well,
they will outgrow others quickly, leaving the norms of the rows of $W$
very unbalanced. Note that this is in particular problematic with the
RPU noise management, since the inputs are divided by $\max |x_i|$ so
that weakly activated inputs get buried in the output noise. If they
are suppressed to such a degree that the output becomes smaller than
the smallest ADC resolution, they may become essentially zero in the
output.

Thus we propose here to counter-act this drive to unbalance rows of
$W$ by using (channel-wise) normalization of the input
(z-scoring). Note that this is very similar to spatial batch
normalization (BN) used by default (for other reasons) in modern deep
learning architectures, such as ResNet~\cite{resnet}. However, since
we here simply want to maintain the variance across the inputs, we do
not train an additional scale or bias per channel, like typically done
in BN~\cite{bn}, and we place the normalization \textit{before} each
layer. Additionally, we z-score the inputs across the batch before a
fully-connected layer, not only convolutions. As in BN, during testing
we use running mean and variances from the train runs and are fixed
during testing.

With this modification of the network structure, where we replace the
LRN with activation z-scoring, we find that the 3-layer CNN even beats
the original model (using LRN) considerably, when trained in software
with floating point accuracy without any RPU hardware simulations or
noise (with identical learning rate and without weight decay), see
\figref{full}~(right, blue curve).

More importantly, the baseline RPU model now performs much more stable,
and at least the model with $4\times$ more states almost reaches FP
performance \figref{full}~(right, red curve). However, the
baseline RPU model with more limiting number of weight states is still 10\%
off the FP reference (green curve).

\subsection{``Virtually'' remap weight ranges to maximize usable
  states}
Our second suggestion is to ``virtually'' remap the weight bounds to
an usable weight range per layer, which not only maximizes the
available physical states, but also has the additional benefit that
saturation at the weight bounds acts as n adequate weight regularization.
The motivation comes from the following observation. 

Between layers of a deep network, it is important to approximately
maintain a 1:1 ratio of the standard deviation of the input and
outputs of a layer. In particular, assume that
$y_j = \sum_{i=1}^nw_{ji}x_i$ and $x_i\sim {\cal N}(0,1)$ behaves like
a standard normal random variable. Let's for simplicity assume that
the matrix $W$ has all identical entries $w$. Then, it is easy to see
that the output standard deviation is of the order of $\sqrt{n}$, i.e.
$y_j\sim{\cal N}(0,w\sqrt{n})$. Thus, the standard deviation of the
output is proportional to the square root of number of dimensions of
the input. This idea, which also holds for more general $W$,  is the basis for all weight initialization
techniques, such as Xavier or He
initialization~\cite{glorot2010understanding,he2015delving}. In
Caffe2's Xavier implementation, the weight is initialized uniformly in
the range
$w\in
\left(-\frac{\sqrt{3}}{\sqrt{n}},+\frac{\sqrt{3}}{\sqrt{n}}\right)$. Note
that the division by $\sqrt{n}$ achieves that the output standard
deviation is roughly of the order of the input standard deviation,
independent of the number of column of the weight matrix.

Such weight initializations were shown to be very essential in successful
training of deep ANNs, as it prevents an explosion of the
activations through the layers and normalizes for different weight
matrix sizes~\cite{glorot2010understanding,he2015delving}.

\begin{figure}[t]
  \centering
  \includegraphics[width=\linewidth,clip,trim=2cm 0 2cm 0]{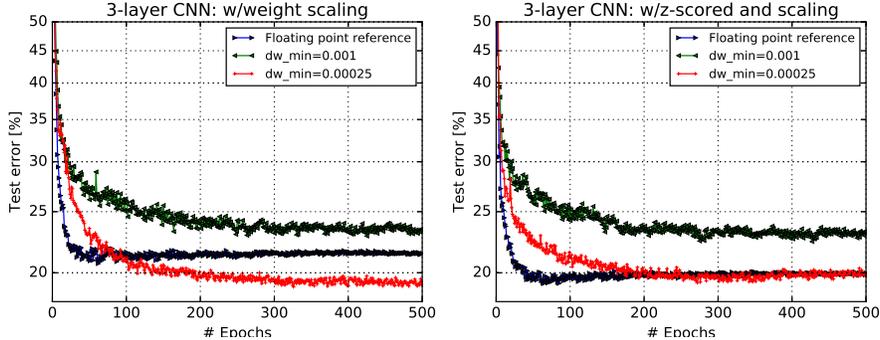}
  \caption{\small Same network and settings as in \figref{full} but
    now additionally applying the described weight scaling. (Left)
    Using the original 3-layer CNN with RPU simulation.  Note that
    proper weight scaling makes the RPU network actually beat the
    performance of the FP performance, when using $4\times$ more
    states (red), and dramatically improve the baseline RPU model with
    more limited number of states.  (Right) Using the z-scored network
    in combination with the weight scaling does not further improve
    the RPU simulations for this network.  Parameters: as in
    \figref{full}, $\gamma=0.4$}
  \label{fig:full_aws}
\end{figure}

\subsubsection{Proper weight scaling for RPUs}
We propose to take advantage of the requirement of scaling the weight
into a smaller range for larger weight matrices. The insight is that,
even for a trained model the requirement of having the similar input
and output standard deviation should still hold\footnote{at least for
  intermediate layers, the final layer before the softmax might be an
  exception}. Thus individual weights should not deviate ``too much''
from their initialization bounds. That means after training it is
still $\max |w_{ij}| = \frac{\beta}{\sqrt{n}}$ with $\beta\approx 1$
or at most a few times larger than that. Given the limited weight
resources in analog space, we thus do not need to waste weight states
to code for weight values $\gg\frac{1}{\sqrt{n}}$. Our approach is
thus to virtual map the weight range
$(-\frac{\beta}{\sqrt{n}}, \frac{\beta}{\sqrt{n}})$ into the original
weight range $(-w_b,w_b)$. This can be achieved in the RPU array by
adjusting the mapping of weight values to resistive values accordingly
without changing the hardware specifications. Or it could be done in
digital, by additional scaling the digital output of the RPU
calculation (of forward and backward passes) by a factor of
$\frac{\beta}{\sqrt{n}w_b}$, to virtually re-scale the weight
range. Note that in this case the learning rate $\lambda$ has to be
divided by the same factor for that particular layer, to re-scale it
properly to the re-mapped weight range.

We use $\beta\equiv \frac{\sqrt{3}}{\gamma}$ (with $\gamma\le 1$) for
all layers if not otherwise stated, and initialize the weights
uniformly in the range $(-\gamma w_b,\gamma w_b)$. Thus we allow the
weights to grow $\frac{1}{\gamma}$ times beyond the maximal
initialization value and therefore maximize the number of available
states in this range. Note that we use the bound management to ensure
that signals are not saturated because of the limited output range
(see above).

We trained the 3-layer CNN again using the above scaling and
normalization approaches. We find that, in particular when the number
of states is more limited (e.g. 1200 for the baseline RPU model), scaling
the weight bounds properly is the most effective to increase
performance (see \figref{full_aws}, left). In this network architecture, the
performance increased by least 10 \%-points. Moreover, the RPU now is
much better regularized (no late increase in test error as
in~\figref{full}~left), as the saturation at the limited weight range
prevents individual weights to become dominating.

The weight scaling approach seems to also normalize the
activation correctly, so that z-scoring does not gain on top of using
the weight scaling (compare \figref{full_aws} right).

Note that when the number of states is increased 4-fold, the RPU model
in fact now beats the floating point reference, despite the noise and
pulsed weight update, because of better regularization properties. In
conclusion, the weight scaling forces the RPU to operate on the
correct weight range and maximizes the available states in this
range. Thus, the requirement for the amount of states is lessened.

\subsection{Larger networks and data sets}
\subsubsection{ResNet on CIFAR}
\begin{figure}[t]
  \centering
  \includegraphics[width=\textwidth,clip,trim=1.5cm 0.6cm 1.5cm 0cm]{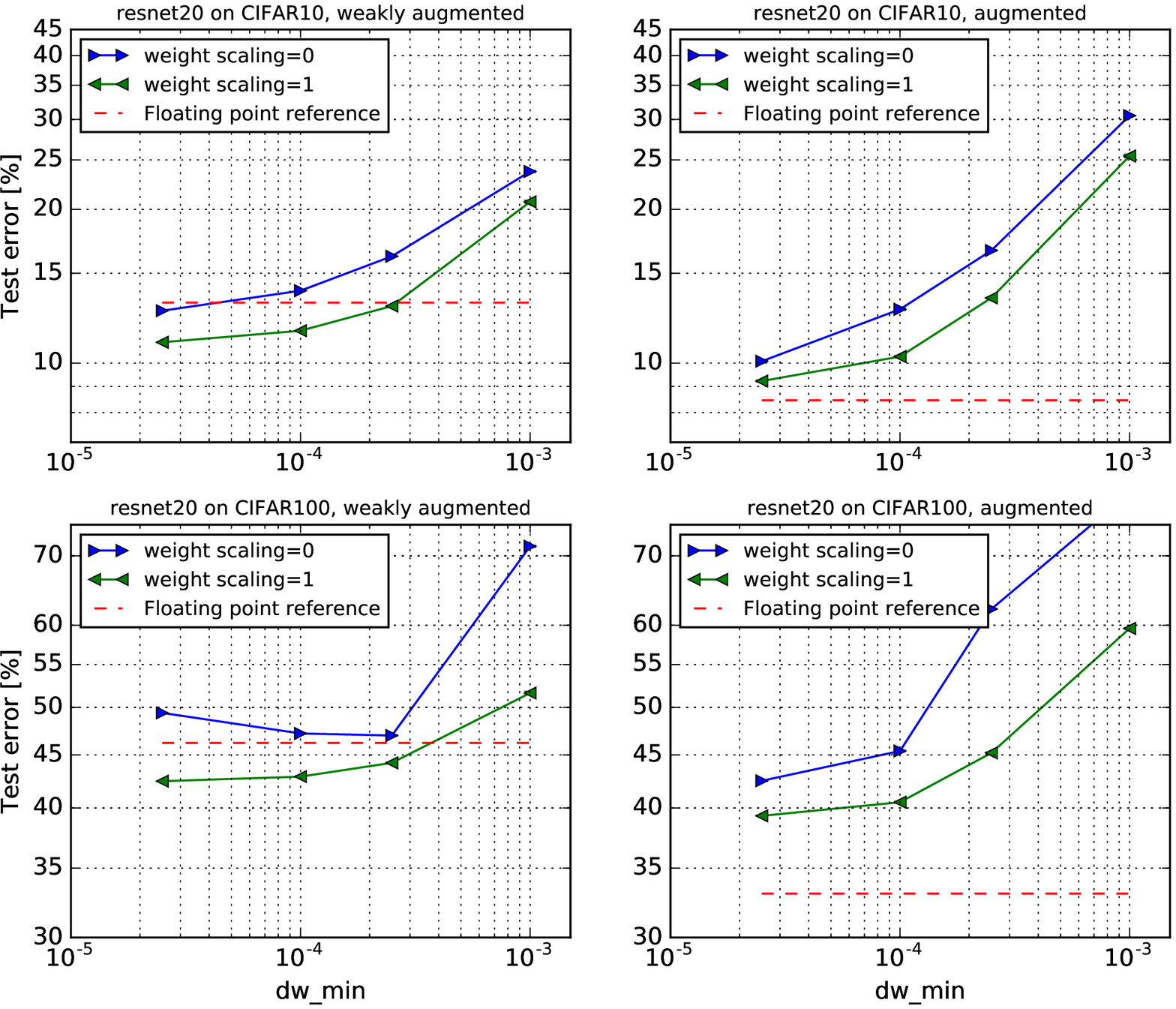} 
  \caption{\small Training on Resnet20 (18 conv layers) on CIFAR10
    (upper plots) and CIFAR100 (lower plots). Left plots uses weak
    augmentation as before (color jitter and random mirroring), where
    as use stronger input augmentation (scale jitter 1.3 times and
    random order shuffeling). Note that 4x times more states than the
    baseline RPU model is enough to beat the floating point reference,
    when using weight scaling. However, while input augmentation
    strongly improves the floating point accuracy, it improves the RPU
    network less well (and actually significantly reduces the
    performance in case of $\dwmin=0.001$), potentially because
    the input becomes too erratic for the noisy pulsed update
    process. Higher number of states, recover the lost performance to
    some degree.  Parameters: best of $\lambda=0.2$ or $\lambda=0.4$,
    multiplied by a factor each 150 epochs so that 1\% of $\lambda$ is
    reached after the 500 epochs training; batch size 100; $\gamma=1$.  }
  \label{fig:resnet}
\end{figure}

We further investigated the number of states needed for training
ResNet20 (i.e. $n=3$ in \cite{resnet}~4.2) on CIFAR10 and CIFAR100
data sets. We use the above weight scaling technique. Note that ResNet
has batch normalization per default which we use here instead of the
z-scoring (results are slightly better for BN in case of ResNet). We
applied weight scaling and varied the number of states of our baseline
RPU model (by changing $\dwmin$) and found that (see
\figref{resnet}) (1) weight scaling improves the performance
considerably, in particular, when the number of states is smaller, (2)
with weak data augmentation, pulsed update RPU generalizes better than
the FP reference, (3) with strong data augmentation (30\% scale
jitter with random cropping, and random image shuffling per epoch) the
RPU model needs at least $48$K states to be able to come close to the
FP reference, which improves dramatically with stronger data
augmentation.

Our results indicates that larger networks and more challenging tasks
(such as CIFAR100) demand more resources in terms of number of
states. In particular, analog noise and finite number of states seem
to limit the performance gain achieved by strong data augmentation,
which is very effective method for improving generalizability for ANNs
trained with floating point accuracy and limited dataset sizes.

\subsubsection{AlexNet  on Imagenet}
For the first time, we simulate analog network training on close to
state-of-the-art scale using pulsed weight update and noisy backward
pass within the specification of analog RPUs. We train
AlexNet~\cite{alexnet} from scratch on the Imagenet database, a
problem, that is more than 40000 times more challenging than training
LeNet on MNIST (in terms of MACs per epoch), which nevertheless is
still used as a typical benchmark for analog hardware
evaluation~\cite{sze2017efficient}.

\begin{figure}[t]
  \centering
  \includegraphics[width=0.5\textwidth]{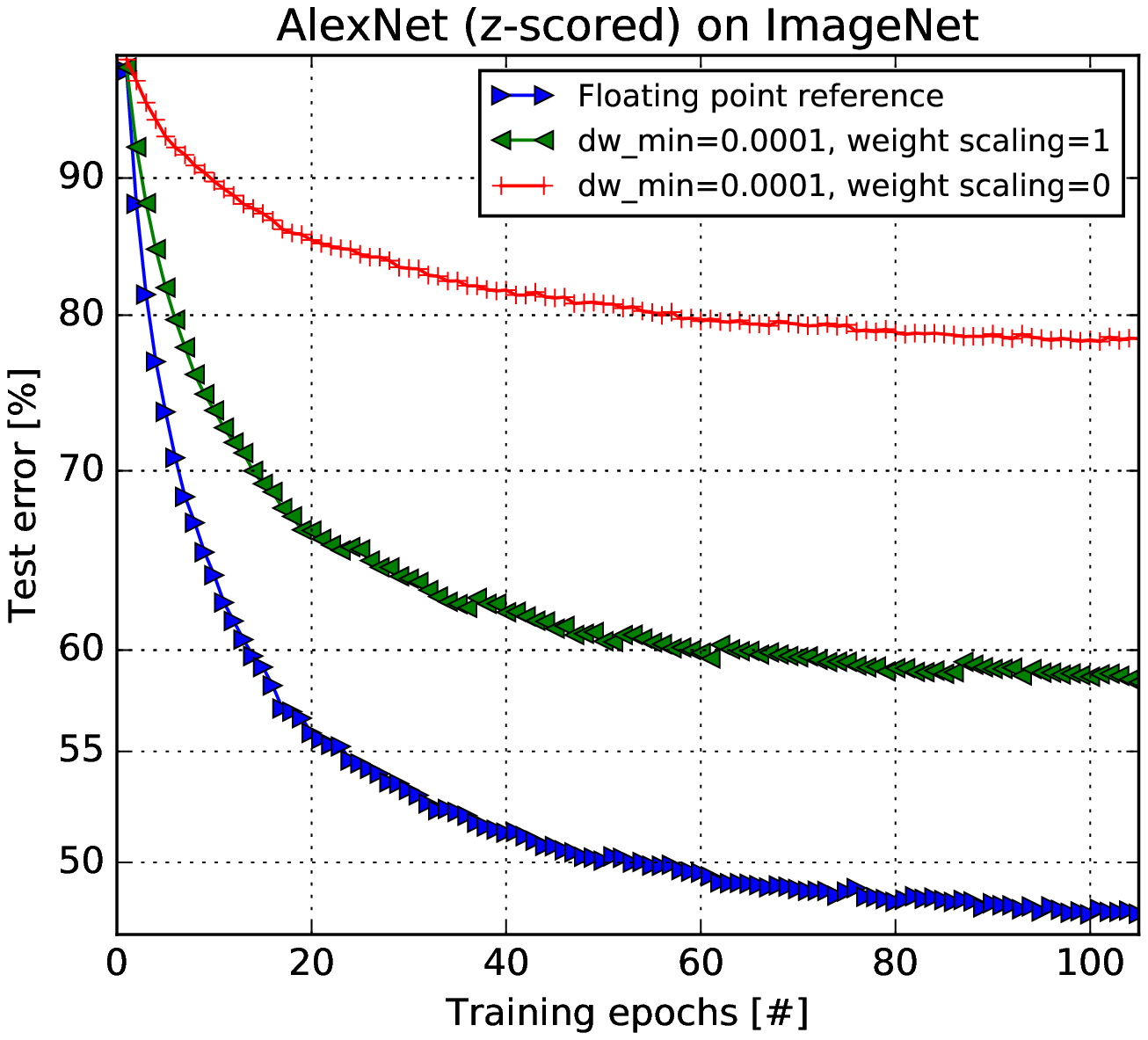}
  \caption{\small Training on AlexNet on ImageNet with (symmetric)
    pulsed update and z-scored activations. Effective number of states
    is 12K, otherwise the baseline RPU model is used as above. Note
    that enabling the weight scaling dramatically improves training
    performance.  However, even with weight scaling, 12K states are
    not enough to reach FP performance. Parameters: $\lambda=0.00125$,
    multiplied by 0.8 every 15 epochs; only weak augmentation; batch
    size 25.  }
\label{fig:alexnet}
\end{figure}

We find that using AlexNet off-the-shelf is not trainable with our
baseline RPU model (even with floating point update but limited
ADC/DAC resolution, not shown). We thus applied the above z-scoring
techniques between each layer and tested the effect of additionally
using weight scaling on the number of device states necessary. In
\figref{alexnet} we show that, using 12K states during pulsed
update ($\dwmin=0.0001$, 10x more states than our baseline RPU
model), reaches top-1 test error of only slightly below 80\%. On the
other hand, enabling weight scaling, we find that the test error is
dramatically improved, by about 20 \%-points, reaching test errors of
slightly below 60\%. That this positive effect of weight remapping is
larger for AlexNet than for ResNet is understandable since the
dimensions of the weight matrices (and therefore the scale term
proportional to $\sqrt{n}$) is much larger (up to 9216) than for
ResNet on CIFAR (up to 577).

However, our results also indicate that for reaching the floating
point accuracy of below 50\%, 12K states are not enough. Thus,
although our approach dramatically improves accuracy of analog
approaches, even with symmetric updates, reaching floating point
accuracy with analog hardware on larger scale networks remains a
challenge and probably requires additional algorithmic improvements
similar to those presented here.

\section{Conclusion}  
Scaling up simulations of analog crossbar approaches for acceleration
of ANN training is a necessary and essential prerequisite for
evaluating and finding new algorithmic or hardware design solutions
that minimize the accuracy gap in respect to the floating point
reference. We show that simple algorithmic modifications, such as proper
normalization and weight range remapping, can dramatically improve
training performance on large-scale ANNs with constraint weight
resources (range and precision).

Our results also highlight the importance of having a digital part
between layers that could accommodate not only the activation
functions and pooling, but also the essential bound and noise managing
techniques, and other algorithmic compensatory measures such as
normalization and weight remapping as suggested here. To maintain the
advantages of the crossbar architecture, these digital operations need
to be computed locally close to the array's peripheral edges. 

While our algorithmic improvements yield a considerable improvement in performance of
training in-memory on analog RPU arrays, the number of required states
to match the FP reference for large scale networks is still beyond
current materials~\cite{haensch2019next}. However, possible solution
pathways exist, e.g. one RPU device might be a combination of multiple
physical devices (of possibly different significance), which could
dramatically enlarge the number of attainable states~(e.g. as in
\cite{shafiee2016isaac,ambrogio2018equivalent}).

Note that the ConvNets evaluated here are generally not well suited
for analog architectures because of the re-use of the kernel matrix,
which slows down computation in analog systems (see
\cite{rasch2018efficient} for a discussion). However,
\cite{rasch2018efficient} also suggested an algorithmic modification
of ConvNets to overcome this problem and better map the ConvNet
architecture to analog RPU systems by replicating kernel matrices and
train them in parallel. How noise and limited number of states are
effected in these so-called RAPA-ConvNets remains to be investigated.

While we here have simulated the training process on RPU arrays, a
related problem is training ANNs in a RPU hardware-aware manner, to
optimize the inference performance on RPU devices. These simulation
would include all components used in our simulations, except that the
weight update and backward pass would be considered perfect and
noise-free, which would dramatically improve the attainable accuracy,
even with much less available states or more noise in the forward
pass. Thus, training in analog space is a much more challenging
problem than training to optimize inference on analog RPUs. 

In summary, we here explored the challenges of analog hardware design
constraints for training large-scale networks and suggested a number
of algorithmic compensatory measures to lessen the performance impacts
of noise, limited number of states and limited weight ranges, even if
the device switching behavior would be ideal, as assumed here. While
we show a dramatic improvement, our results also suggest that more
concentrated research efforts on algorithmic, material, and
system-level are needed to be able reach state-of-the-art training
performance of analog ANN accelerators.

\bibliographystyle{abbrv}
\bibliography{pulsed}

\end{document}